\documentclass[11pt]{article}

\usepackage[margin=1in]{geometry}
\usepackage{lmodern}
\usepackage[T1]{fontenc}
\usepackage{amsmath, amssymb}
\usepackage{booktabs}
\usepackage{graphicx}
\usepackage{microtype}
\usepackage[numbers]{natbib}
\usepackage[colorlinks=true, linkcolor=blue, citecolor=blue, urlcolor=blue]{hyperref}
\usepackage{xspace}

\newcommand{\At}{\ensuremath{\hat{A}_t}\xspace}

\title{Re-feeding Is Not Replaying:\\
Measuring Replay Noise in\\
Counterfactual Token-Credit Estimation}

\author{
  Nils Matteson\\
  thaw \,$\cdot$\, Northeastern University\\
  \texttt{matteson.ni@northeastern.edu}
}

\date{June 2026}

\begin{document}
\maketitle

\begin{abstract}
Per-token counterfactual credit estimation asks which token in a language-model
rollout caused the final answer to be right or wrong: cut the transcript at a
pivot, substitute an alternative token, replay continuations, and compare
outcomes. Published methods implement the replay by \emph{re-feeding} the
transcript prefix as a fresh prompt, assuming the re-fed prefix reproduces the
state the model passed through during generation. We measure what that
assumption costs on a stock inference engine, with a three-pass design:
continuations resumed from the verified decode-time KV state, an identical
second exact pass serving as a replica noise floor, and a re-feed pass.
Across six configurations and three models (including a GRPO-trained
checkpoint), at low-margin decision tokens, re-feeding changes the credit
estimate at rates 14--28 percentage points above the replica floor
(7--21pp under a treatment-independent conditioning; problem-clustered
$t = 2.9$--$6.4$). Most changes are zero-boundary
crossings of the quantized estimator rather than polarity reversals, and the
perturbation is consistent with mean-zero, so averaged quantities are
largely safe; but
selection is not: a critical-token set chosen by thresholding $|\At|$ under
re-feed overlaps the exact-resume selection at Jaccard 0.34--0.90, versus a
0.63--0.96 replica ceiling. A causal confirmation closes the loop: rerunning
the harness under vLLM's batch-invariant kernels makes all three passes
identical on every measured channel, with both disagreement rates exactly
zero. Even the floor
is informative: replica passes disagree on 9--23\% of eligible estimates,
so single-sample credit measurements at decision tokens are unreliable under
\emph{any} replay. Settings were fixed in advance with a locked decision
rule; cache hits underlying the second campaign's ``exact'' passes are
instrumented (100\% hit
rate, 3{,}434 pivots; 3{,}829 including the batch-invariant run); total compute was under \$10. We recommend that
counterfactual credit studies resume decoder state or use batch-invariant
kernels, and report a replica floor.
\end{abstract}

\section{Introduction}

Reinforcement learning with verifiable rewards has made per-token credit
assignment a central measurement problem for language-model reasoning. Group
relative policy optimization (GRPO) \citep{shao2024deepseekmath} and its descendants
train on sequence-level rewards, so a growing line of work asks the
finer-grained question: which individual tokens in a rollout caused success
or failure? The standard answer is counterfactual replay. VinePPO
\citep{kazemnejad2024vineppo} states the recipe's premise directly: language
environments ``allow us to reset directly to any intermediate state simply by
re-feeding the partial context.'' Critical-token methods
\citep{lin2024critical, ruan2025cft} identify decisive tokens by regenerating
continuations from token positions; causal credit constructions
\citep{khandoga2026beyond} and exact-counterfactual agent-credit methods
\citep{chen2026exact} likewise reconstruct intermediate state from text.

The shared mechanical assumption is that re-feeding tokens $x_{1..t}$ puts
the model in the state it was in when it generated token $t$. At the bit
level it does not. Floating-point reduction order inside attention and
matmul kernels depends on batch composition and on how a sequence is split
between prefill and decode \citep{he2025nondeterminism}, so a re-fed prefix
yields slightly different logits than the live decode state did. The working
assumption in the credit literature is that these differences wash out of
the final estimate. C3 \citep{chen2026exact} asserts the premise explicitly:
that any decision point can be restored exactly from text. This paper
measures that premise on the engines the literature actually runs on, and
quantifies precisely what it costs, what it does not cost, and what
eliminates it.

\paragraph{Contributions.}
\begin{enumerate}
  \item \textbf{A three-pass audit design with a replica floor.} We compare
  credit estimates from (i) continuations resumed from the decode-time KV
  state (cache hits instrumented per request: 100\% verified across 3{,}434
  pivots in the five second-campaign stock-engine runs, two of which compose
  the pooled configuration; 3{,}829 including the
  batch-invariant run), (ii) an identical second exact pass, and (iii) a re-feed pass.
  Pass (ii) converts ``how often do estimates disagree?'' into ``how much
  disagreement is attributable to the re-feed procedure \emph{beyond}
  replica noise?'' The design runs on stock vLLM \citep{kwon2023vllm} with
  no kernel modifications.
  \item \textbf{Quantification with disclosed decomposition.} At low-margin
  decision tokens, re-feeding changes the credit estimate at rates 14--28pp
  above the
  floor across six configurations and three models (clustered
  $t=2.9$--$6.4$). We decompose the changes: most are zero-boundary
  crossings of the $1/K$-quantized estimator; strict polarity reversals are
  0.5--6.2\% under re-feed versus 0--5.9\% at the floor. The perturbation is
  consistent with mean-zero ($t = -1.2$), so aggregated estimates are
  largely unaffected.
  \item \textbf{A downstream consequence: selection.}
  Threshold-based critical-token selection, the construction of
  \citet{lin2024critical} and \citet{ruan2025cft}, selects a materially
  different token set under re-feed: Jaccard overlap with the exact-resume
  selection is 0.34--0.90 across configurations, against a replica ceiling
  of 0.63--0.96.
  \item \textbf{A causal confirmation.} Rerunning the identical harness
  under vLLM's batch-invariant kernels (\texttt{VLLM\_BATCH\_INVARIANT=1},
  implementing the fix of \citet{he2025nondeterminism}) makes all three passes
  identical on every stored channel (credit estimates, greedy probe
  sequences, probe logprobs): both disagreement rates collapse to exactly
  zero and 100\% of probes are bit-exact. In the tested configuration the
  effect is batch-variant kernel numerics, and the published fix
  eliminates it.
  \item \textbf{The floor as a refinement.} Prior work established that
  batching numerics shift benchmark accuracy and flip greedy decoding
  \citep{yuan2025numerical, song2024greedy}. We measure the same hazard at
  the unit relevant to credit assignment: two bit-identical request batches
  disagree on 9--23\% of eligible per-pivot estimates. Single-sample credit
  measurement at decision tokens is unreliable under any replay method.
\end{enumerate}

Experimental settings (temperature, $K$, grading, pivot filter) and the
decision rule were fixed before any GPU run, with one dated amendment
(a difficulty screen) added after a smoke run and before any full-run
results; the registration is self-attested (published in the repository,
without an external timestamp) and the paper relies on it only as
disclosure, not as proof. Total compute across both experimental campaigns
was under \$10 on rented A100s. All per-pivot records, logs, the harness,
and the analysis function that emits every number in this paper are public.

\section{Related Work}

\paragraph{Counterfactual token credit.} VinePPO
\citep{kazemnejad2024vineppo} builds Monte Carlo value estimates on the
premise that re-feeding partial context resets the environment state.
\citet{lin2024critical} identify critical tokens by rollouts from token
positions; critical-token fine-tuning \citep{ruan2025cft} forces
alternative tokens and regenerates continuations. \citet{khandoga2026beyond}
construct causal per-token credit by masking reasoning spans and measuring
answer-probability changes. C3 \citep{chen2026exact} shows exact
counterfactuals beat approximate ones for credit assignment among
cooperative LLM agents, at the granularity of whole agent actions, and
premises its method on text restoration being exact. All reconstruct
intermediate state from text; none resume the original decoder state as a
reference, and none report a replica control bounding the replay procedure's
own contribution. A contemporaneous survey of the area
\citep{zhang2026survey} repeats the premise: generating rollouts from any
intermediate prefix is treated as trivially available, with no discussion
of replay fidelity. Separately, tree-structured rollout methods
\citep{li2025treepo} do resume live KV state at branch points, but as an
efficiency mechanism during training, without comparing resume to re-feed
or auditing exactness; our scope claim is therefore about credit
\emph{measurement} studies, not about all systems that branch from cached
state. \citet{chatzi2024counterfactual} study counterfactual token
generation via shared sampling noise (Gumbel-max), an orthogonal notion of
counterfactual that fixes randomness rather than internal state.

\paragraph{Inference nondeterminism and training mismatch.}
\citet{he2025nondeterminism} identify the lack of batch invariance as the
root cause of
LLM inference nondeterminism and eliminate it with batch-invariant kernels; SGLang
ships a comparable deterministic mode \citep{sglang2025deterministic}.
\citet{yuan2025numerical} quantify accuracy swings and greedy argmax flips
from batch-size numerics; \citet{song2024greedy} show benchmark evaluation
should not ignore sampling and numeric nondeterminism. On the training side,
\citet{zhong2026tim} diagnose training-inference mismatch (TIM: rollout engine
and trainer disagreeing on logprobs) as an independent cause of RL collapse,
and \citet{qi2025fp16} show FP16 precision largely removes that mismatch.
These lines concern the serving endpoint and the training loop. Our subject
is different: the \emph{measurement methodology} of the credit-assignment
literature, where the comparison pair is re-fed prefill state versus the
original decode-time state, on stock engines. The lines compose: He et
al.\ built the fix, the TIM line covers training, and this paper covers
measurement, including a direct experimental confirmation that the
batch-invariant fix closes the measurement gap (Section~\ref{sec:batchinv}).

\section{Method}

\subsection{Setup and estimator}

For each problem we sample $N = 8$ trunk rollouts at temperature $0.7$ with
top-$k$ logprobs recorded. The engine caches KV blocks as it decodes, so the
state every rollout passed through remains resident in the prefix cache at
block granularity. A \emph{pivot} is a generated token whose top-1/top-2
logprob gap is below $1.0$ (a low-margin decision token, the population
that credit methods target), restricted to positions within 2 tokens past a
16-token block boundary (block size read from the engine configuration and recorded), at
most 5 per rollout, earliest first.

At pivot position $t$ with sampled token $a$ and best alternative $a'$, the
credit estimate under a given replay method is
\[
\At \;=\; \frac{1}{K}\sum_{k=1}^{K} R\big(x_{1..t-1},\,a,\,c_k\big)
\;-\;
\frac{1}{K}\sum_{k=1}^{K} R\big(x_{1..t-1},\,a',\,c'_k\big),
\]
where $c_k, c'_k$ are sampled continuations and $R$ grades the completed
solution by exact match on the GSM8K \citep{cobbe2021gsm8k} final answer.
Continuation seeds are a deterministic hash of (problem, rollout, pivot,
arm, $k$), so all passes receive identical sampling randomness.

\subsection{Three passes}
\label{sec:passes}

\begin{description}
  \item[Exact pass 1.] All continuation requests for a problem are submitted
  while the trunk's decode-time KV blocks are resident. Each request's
  prompt cache-hits the state the rollout actually traversed. This is
  \emph{instrumented}, not assumed: the engine's per-request
  \texttt{num\_cached\_tokens} is recorded and checked against the
  block-aligned pivot position. Across all second-campaign runs, 100\% of
  3{,}829 pivots verified on both exact passes.
  \item[Exact pass 2 (replica floor).] The identical request batch is
  submitted again, cache warm. Disagreement between passes 1 and 2 is the
  noise floor of the measurement itself. One impurity is disclosed: pass 1's
  continuations also populated the cache, so pass 2's requests can hit
  slightly more cached state than pass 1's did, and its step-level batch
  composition differs accordingly. This makes the floor, if anything,
  slightly conservative as a baseline for the re-feed comparison.
  \item[Re-feed pass.] The prefix cache is reset (the reset's return value
  is checked; a silent failure would convert this pass into a third exact
  pass) and the identical batch is submitted. The trunk is freshly
  prefilled, exactly as in published credit pipelines. We attribute
  differences beyond the floor to \emph{the re-feed procedure as actually
  run}: that includes both the recomputed prefix state and the changed
  step-level batch dynamics that heavy chunked prefill induces, because real
  re-feed pipelines batch their requests the same way.
\end{description}

The primary metric is the fraction of eligible pivots at which
$\mathrm{sign}(\At)$ under re-feed differs from exact pass 1, against the
same quantity for exact pass 2. Eligibility is reported under two
conditionings: the pre-specified metric (any pass nonzero; shared denominator)
and a treatment-independent conditioning ($\At^{\text{exact1}} \neq 0$),
since the former lets the re-feed pass create its own eligibility (77 of 636
pooled eligible pivots are nonzero only under re-feed, versus 18 only under
the replica). Sign changes are decomposed into zero-boundary crossings
(one side zero; ``zero-crossings'') and strict polarity reversals. Each pass also issues one
greedy 32-token probe per arm with logprobs recorded, giving a grading-free
channel; probe deltas are per-pivot maxima over the two arms.

\subsection{Why decode-time resume is the reference}

The decode-time KV state is the state the model was actually in when it
chose token $t$; a counterfactual \emph{do}-operation at $t$ is defined
relative to it. Resuming it requires no kernel modifications on
paged-attention engines. Two scope notes. First, ``exact'' is exact to the
16-token block boundary: the trailing 0--2 tokens plus the forced token are
computed as a short prefill chunk in both exact passes, so the resident KV
is exact while the first post-pivot logits still traverse a fresh kernel
path. Second, the replica floor measures the reference's self-consistency,
not its fidelity to the original decode-step logits; 61--94\% of resumed
probes reproduce the post-pivot logprob bit-exactly across configurations
(versus 2.6--5.7\% for re-feed), which validates self-consistency, and the
batch-invariant confirmation (Section~\ref{sec:batchinv}) closes the
remaining gap by eliminating path dependence entirely.

\subsection{Registration, campaigns, and disclosure}

Temperature, $K$, grading, the pivot filter, and the decision rule were
fixed in a registration document before any GPU run, with one dated
amendment (a trunk-accuracy difficulty screen, blind to drift direction)
added after a smoke run. One mechanism deviation is flagged: the
registration described the exact arm as a snapshot-tool fork of the live
KV state; the implementation uses the engine-native prefix-cache resume
instead (functionally the same decode-time blocks, with no third-party
component). The decomposition, conditioning, clustered-$t$, selection, and
batch-invariant analyses are post-registration additions, labeled as
such. The registration is published in the repository with an honest
provenance note: it is self-attested, without an external timestamp. Two experimental campaigns are reported. The first campaign's
two pooled cohorts were later found to overlap on four problems (the
screened cohort rescanned from index 0); all first-campaign pooled numbers
in this paper are therefore deduplicated, and the second campaign separates
cohort windows structurally and adds the cache instrumentation. Two
first-campaign runs (greedy continuations; the GRPO checkpoint) are
internally valid and reported as such, marked $\diamond$ in
Table~\ref{tab:main}.

\section{Experimental Setup}

\textbf{Models.} Qwen2.5-7B-Instruct \citep{qwen2025};
Qwen-2.5-7B-SimpleRL-Zoo \citep{zeng2025simplerlzoo}, a GRPO-trained
checkpoint from the Qwen2.5-7B base; and Phi-4-mini-instruct
\citep{microsoft2025phi} (a distinct
architecture and tokenizer family). \textbf{Engine.} Stock vLLM 0.22.1
\citep{kwon2023vllm}, bfloat16, prefix caching enabled. The batch-invariant
confirmation sets \texttt{VLLM\_BATCH\_INVARIANT=1} on the same engine.
\textbf{Hardware.} One A100 SXM 80GB per run. \textbf{Data.} GSM8K test
split \citep{cobbe2021gsm8k}; the main pooled configuration uses problems
0--19 plus 20 mixed-difficulty problems screened from index 20 upward (no
overlap by construction). \textbf{Cost.} Under \$10 total across both
campaigns.

\section{Results}
\label{sec:results}

\begin{table}[t]
\centering
\small
\setlength{\tabcolsep}{4.5pt}
\begin{tabular}{llrrrrrr}
\toprule
Configuration & Model & $n$ & Re-feed (pol.) & Floor (pol.) & Excess & $t_{\text{clust}}$ & $z$ \\
\midrule
Main pooled ($t{=}.7$, $K{=}4$) & Qwen2.5-7B-Inst & 636 & 33.8\% (3.9) & 15.1\% (1.1) & $+18.7$ & 5.51 & 8.61 \\
$K{=}8$                & Qwen2.5-7B-Inst & 490 & 35.7\% (4.9) & 10.4\% (1.0) & $+25.3$ & 6.36 & 9.68 \\
$K{=}16$ (10 prob.)    & Qwen2.5-7B-Inst & 239 & 36.8\% (5.9) & 23.0\% (5.9) & $+13.8$ & 2.87 & 3.71 \\
Greedy cont.\ ($t{=}0$, $K{=}1$)$^{\diamond}$ & Qwen2.5-7B-Inst & 189 & 40.2\% (0.5) & 13.8\% (0.0) & $+26.5$ & 3.64 & 5.66 \\
GRPO ckpt ($t{=}.7$, $K{=}4$)$^{\diamond}$ & SimpleRL-Zoo-7B & 373 & 35.4\% (3.8) & 9.1\% (1.1) & $+26.3$ & 5.50 & 8.66 \\
Phi-4-mini ($t{=}.7$, $K{=}4$)  & Phi-4-mini      & 370 & 48.6\% (6.2) & 20.8\% (3.0) & $+27.8$ & 3.47 & 7.79 \\
\midrule
Batch-invariant kernels & Qwen2.5-7B-Inst & 106 & \textbf{0.0\%} & \textbf{0.0\%} & 0 & --- & --- \\
\bottomrule
\end{tabular}
\caption{Estimate disagreement with exact resume on each run's common
eligible set: total rate, with strict polarity reversals in parentheses
(percent). Excess in percentage points, computed before rounding.
$t_{\text{clust}}$ is a
problem-clustered paired $t$ over per-problem rate differences; $z$ is the
(unclustered) McNemar statistic, reported for comparability.
$^{\diamond}$First campaign, prior to cache instrumentation; design
identical (second-campaign instrumentation verified 100\% cache-hit rates).
In the greedy row, trunks are sampled at $t{=}0.7$ and only continuations are
greedy.
Under batch-invariant kernels all three passes are bitwise identical.}
\label{tab:main}
\end{table}

\subsection{Re-feed changes estimates far above the floor; most changes are
zero-crossings}

Table~\ref{tab:main} gives the primary result. Re-feeding disagrees with the
exact-resume estimate at 14--28 percentage points above the replica floor in
every stock-engine configuration, with problem-clustered $t$ between 2.9 and
6.4 (per-problem sign splits run 29-worse-versus-5 in the pooled
configuration, 19-versus-0 at $K{=}8$, 16-versus-2 on the GRPO checkpoint).
Under the treatment-independent conditioning the excess is smaller but
decisive everywhere: $+11.7$pp pooled, $+19.0$ at $K{=}8$, $+7.4$ at
$K{=}16$, $+11.9$ greedy, $+17.1$ GRPO, $+21.1$ Phi. There is no monotone
$K$ trend: $K{=}8$ is the strongest of the three $K$ settings.

\begin{figure}[t]
\centering
\includegraphics[width=\linewidth]{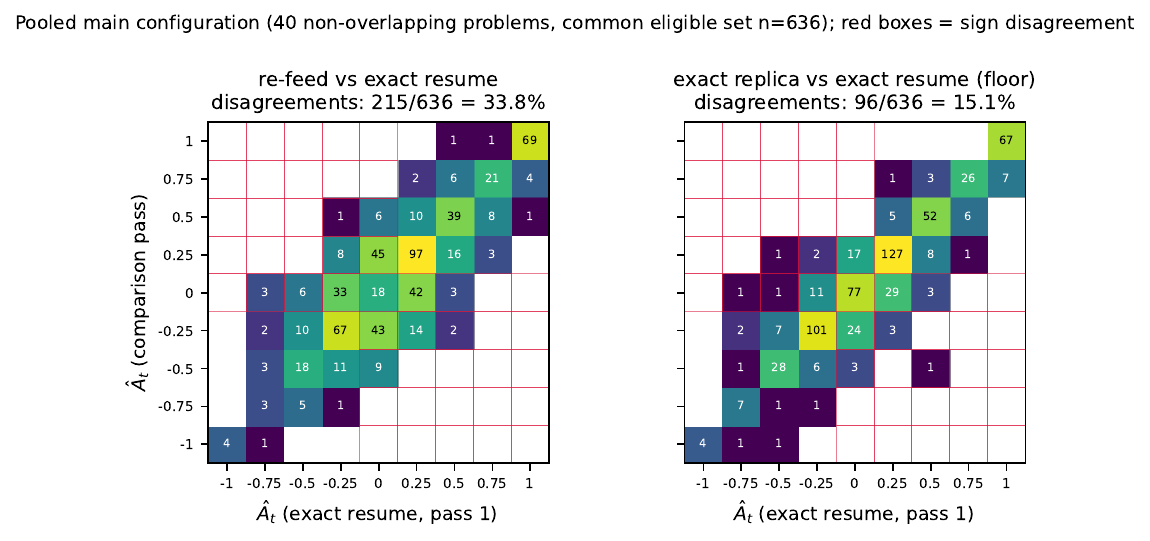}
\caption{Joint distribution of \At under exact resume (x-axis) and the
comparison pass (y-axis), pooled main configuration. Red-outlined cells
disagree in sign with exact pass 1; most disagreement mass sits in the
zero row/column rather than in opposite-sign quadrants.}
\label{fig:quadrants}
\end{figure}

\begin{figure}[t]
\centering
\includegraphics[width=0.95\linewidth]{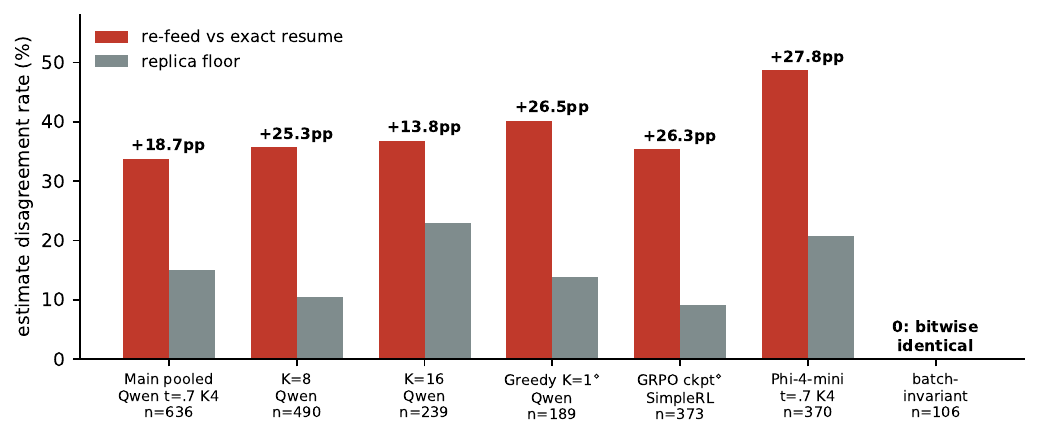}
\caption{Disagreement rate of re-feed (red) versus the replica floor (gray)
across all configurations. The excess survives the removal of sampling
noise (greedy), grows on a second model family (Phi-4-mini), holds on a
GRPO-trained checkpoint, and collapses to exactly zero under
batch-invariant kernels.}
\label{fig:configs}
\end{figure}

The decomposition matters and is disclosed in the same table: strict
polarity reversals (positive credit becoming negative or vice versa) are
0.5--6.2\% under re-feed against 0--5.9\% floors; most disagreements are
the quantized estimator crossing zero. The perturbation is consistent with mean-zero
(pooled bias $-0.011$, $t=-1.23$, 95\% CI $[-0.029, +0.006]$), so quantities that average many pivots
are largely insulated. The effect is also flat in pivot margin (pooled:
34.2\% below a 0.3 logprob gap versus 33.6\% above; floors 12.0\% and
16.9\%). Figure~\ref{fig:quadrants} shows the full joint distribution;
Figure~\ref{fig:configs} compares all configurations.

\subsection{The downstream consequence: critical-token selection changes}

Zero-mean noise still corrupts \emph{selection}. Critical-token methods
\citep{lin2024critical, ruan2025cft} threshold per-token effects to choose
which tokens to train on or analyze. Emulating that construction (select
pivots with $|\At| \geq 0.5$), we find the set selected under re-feed overlaps the
exact-resume selection at Jaccard 0.68 (pooled), 0.68 ($K{=}8$), 0.90
($K{=}16$), 0.59 (greedy), 0.66 (GRPO), and 0.34 (Phi-4-mini), against
replica ceilings of 0.84, 0.87, 0.96, 0.83, 0.88, and 0.63 respectively.
Under batch-invariant kernels both overlaps are exactly 1.0. Roughly
two-fifths to
two-thirds of the gap between a method's selected token set and its own
noise ceiling is attributable to the re-feed procedure.

To be precise about the mechanism: in every selection disagreement we
observe, at least one of the two estimates lies
within one grading quantum ($1/K$) of the threshold, and the re-feed
gap narrows as the estimator gets finer (Jaccard 0.68 at $K{=}4$ to 0.90 at
$K{=}16$). This is not a separate phenomenon from the zero-crossing noise;
it is that noise expressed at the cut. The point is that threshold
selection at the $K$ budgets the literature actually operates at lives
exactly at that cut, and the replica ceilings show such selection is
substantially irreproducible under \emph{any} replay at these budgets,
with re-feeding consuming a further two-fifths to two-thirds of the
remaining
headroom.

\subsection{Bit-level account}

\begin{figure}[t]
\centering
\includegraphics[width=0.72\linewidth]{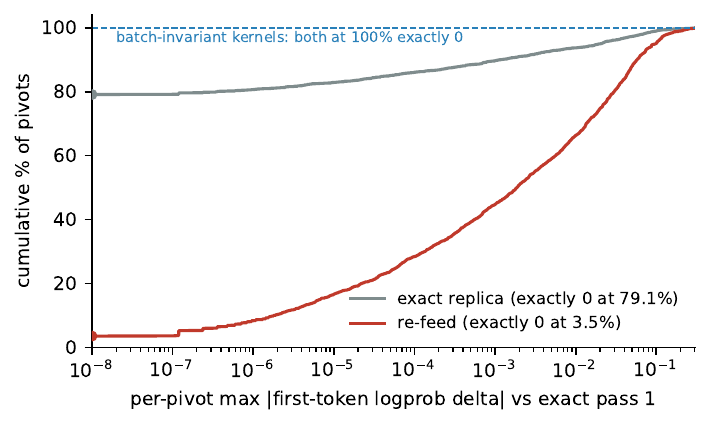}
\caption{Cumulative distribution of the per-pivot maximum first-token
logprob delta against exact pass 1, over four instrumented
second-campaign runs (the two pooled-main cohorts, $K{=}8$, and
Phi-4-mini). The replica's mass concentrates at exactly zero;
re-feed places essentially no mass there. Under batch-invariant kernels both
curves degenerate to 100\% at zero.}
\label{fig:ecdf}
\end{figure}

On the grading-free channel (greedy 32-token probes, per-pivot max over the
two arms), resuming the decode-time state reproduces the first post-pivot
token's logprob exactly (delta $=0.0$) at 61--94\% of pivots across
configurations, while re-feeding does so at 2.6--5.7\%; pooled 90th-percentile
deltas are $3.7\times10^{-3}$ (replica) versus $4.8\times10^{-2}$
(re-feed); Figure~\ref{fig:ecdf} shows the distributions. Probe sequences
diverge within 32 greedy tokens at 22.8\% of
pooled pivots under re-feed versus 7.9\% for the replica.

\subsection{Causal confirmation: batch-invariant kernels eliminate the
effect}
\label{sec:batchinv}

If the measured excess is batch-variant kernel numerics and nothing else,
then under kernels whose reduction order is independent of batch
composition and prefill/decode split \citep{he2025nondeterminism}, all
three passes should be identical. We reran the harness with
\texttt{VLLM\_BATCH\_INVARIANT=1} on the same engine and model at the main
settings ($t{=}0.7$, $K{=}4$), over the first 10 problems of the primary
cohort (395 pivots, 106 eligible). The prediction held exactly: zero
disagreements in both comparisons, 100\% bit-exact probes in both
comparisons, identical estimates on all 395 pivots, and critical-token
Jaccard 1.0 (identity verified on every stored channel: estimates, greedy
probe token sequences, and probe logprobs). This simultaneously closes the
causal attribution and validates the instrument end-to-end: a harness
artifact that manufactured drift would not measure exactly zero when the
numerics are pinned.

\subsection{The floor is also a finding}

Two bit-identical request batches, submitted seconds apart with identical
seeds, disagree on 9--23\% of eligible per-pivot estimates and diverge on
2--14\% of greedy probes. Prior work measured this hazard at the benchmark
level \citep{yuan2025numerical, song2024greedy}; at the per-decision-token
unit relevant to credit assignment it is large enough that single-sample
estimates at low-margin tokens are uninterpretable under any replay method
on stock engines. Any credit-assignment study operating at this granularity
needs a replica floor to be interpretable at all.

\section{Discussion}

\paragraph{What this does and does not show.} It shows that the re-feed
shortcut adds measurement noise far above the replica floor at decision
tokens on stock engines, that this materially changes threshold-based
token selection, and that batch-invariant kernels eliminate it entirely.
It does not show that published aggregate conclusions are wrong: the
perturbation is consistent with mean-zero, and methods that average over
many tokens and
rollouts are largely insulated. It does not claim novelty for the existence
of batch numerics effects \citep{he2025nondeterminism, yuan2025numerical}.
One further scope note: same-process pipelines that re-feed with prefix
caching enabled may silently cache-hit the decode-time blocks and thereby
run the exact arm without knowing it; the indictment here applies to
cache-cold re-feeds, which is what fresh-process and batched-analysis
pipelines do.

\paragraph{Recommendations.} (1) Run counterfactual replay on
batch-invariant or deterministic kernels where available
\citep{he2025nondeterminism, sglang2025deterministic}; our confirmation run
shows this closes the gap exactly, at some throughput cost. (2) Otherwise,
resume the decode-time state: on paged-attention engines this is
engine-native within a process via the prefix cache, and durable
session-snapshot tooling can extend it across processes. (3) In all cases
report a replica floor; it costs one extra batch and is the difference
between an interpretable and an uninterpretable single-pivot measurement.
(4) Fix estimator settings in advance: the disagreement rate is sensitive
to temperature, $K$, grading, and the pivot filter.

\paragraph{Conflict of interest.} The author builds thaw, an open-source
session-snapshot tool for vLLM whose premise (durable exact decoder state)
is adjacent to recommendation (2). The experiments in this paper use stock
vLLM only; the in-process resume used as the exact reference requires no
thaw component, and recommendation (1), a third party's fix, is presented
as the preferred path where available.

\paragraph{Limitations.} One task family (GSM8K); one engine version (vLLM
0.22.1) and GPU type (A100); pivots are conditioned on low margin and
block-boundary proximity, so all rates are statements about decision
tokens, not all tokens; $K \leq 16$ with binary grading keeps the estimator
coarse (the floor quantifies the consequence); the registration is
self-attested; and two first-campaign runs predate the cache
instrumentation (marked in Table~\ref{tab:main}). Passes always run in a fixed
order (exact, replica, re-feed) and there is no re-feed-versus-re-feed
replica, so the re-feed arm's internal variance is unmeasured; we also do
not isolate the recomputed-prefix mechanism from the changed batch
dynamics with a serialized one-request-at-a-time control, which is why the
treatment is defined as the re-feed procedure as actually run. The GRPO
checkpoint was run only in the first campaign. The batch-invariant
confirmation covers one configuration; replicating it across the full grid
is mechanical.

\section{Reproducibility}

The registration (with its dated amendment and provenance note), the
harness (one Python file against stock vLLM), every per-pivot record of all
runs in both campaigns, run logs, and the analysis function that emits
every number and figure in this paper are public in the thaw repository
(\url{https://github.com/thaw-ai/thaw}: \texttt{benchmarks/},
\texttt{paper/refeed-drift/}). Total compute: under \$10 of rented A100
time. The first campaign's cohort-overlap error is documented in the
repository history rather than erased; pooled first-campaign numbers are
deduplicated wherever cited.

\bibliographystyle{plainnat}
\bibliography{refs}

\end{document}